\documentclass[conference]{IEEEtran}
\usepackage{cite}
\usepackage{amsmath,amssymb,amsfonts}
\usepackage{graphicx}
\usepackage{subcaption}
\usepackage{textcomp}
\usepackage{xcolor}
\usepackage[utf8]{inputenc}
\usepackage[T1]{fontenc}
\usepackage{multirow}
\usepackage{amsmath}
\RequirePackage[table,figure]{totalcount}
\usepackage{xcolor}
\usepackage{algorithm}
\usepackage{algpseudocode}
\usepackage{enumitem}
\usepackage{caption}
\usepackage{fancyhdr}

\begin{document}


\title{BanglaMM-Disaster: A Multimodal Transformer-Based Deep Learning Framework for Multiclass Disaster Classification in Bangla}

\author{\IEEEauthorblockN{Ariful Islam}
\IEEEauthorblockA{\textit{Department of Computer Science and Engineering} \\
\textit{Chittagong University of Engineering and Technology}\\
Pahartoli, Raozan-4349, Chittagong, Bangladesh \\
arifulislamnayem11@gmail.com}
\and
\IEEEauthorblockN{Md Rifat Hossen}
\IEEEauthorblockA{\textit{Department of Computer Science and Engineering} \\
\textit{Chittagong University of Engineering and Technology}\\
Pahartoli, Raozan-4349, Chittagong, Bangladesh \\
rifat8851@gmail.com}
\and
\IEEEauthorblockN{Md. Mahmudul Arif}
\IEEEauthorblockA{\textit{Department of Electronics and Telecommunication Engineering} \\
\textit{Chittagong University of Engineering and Technology}\\
Pahartoli, Raozan-4349, Chittagong, Bangladesh \\
mdmahmudularif896@gmail.com}
\and
\IEEEauthorblockN{Abdullah Al Noman}
\IEEEauthorblockA{\textit{Wilmington University} \\
320 N Dupont Hwy, New Castle, DE 19720 \\
anoman001@my.wilmu.edu}
\and

\IEEEauthorblockN{Md Arifur Rahman}
\IEEEauthorblockA{\textit{College of Graduate and Professional Studies} \\
\textit{Trine University}\\
1 University Ave, Angola, IN 46703 \\
Mrahman22@my.trine.edu}
}

\maketitle

\begin{abstract}
Natural disasters remain a major challenge for Bangladesh, so real-time monitoring and quick response systems are essential. In this study, we present BanglaMM-Disaster, an end-to-end deep learning-based multimodal framework for disaster classification in Bangla, using both textual and visual data from social media. We constructed a new dataset of 5,037 Bangla social media posts, each consisting of a caption and a corresponding image, annotated into one of nine disaster-related categories. The proposed model integrates transformer-based text encoders, including BanglaBERT, mBERT, and XLM-RoBERTa, with CNN backbones such as ResNet50, DenseNet169, and MobileNetV2, to process the two modalities. Using early fusion, the best model achieves 83.76\% accuracy. This surpasses the best text-only baseline by 3.84\% and the image-only baseline by 16.91\%. Our analysis also shows reduced misclassification across all classes, with noticeable improvements for ambiguous examples. This work fills a key gap in Bangla multimodal disaster analysis and demonstrates the benefits of combining multiple data types for real-time disaster response in low-resource settings.
\end{abstract}
\begin{IEEEkeywords}
multimodal deep learning,
transformer models,
Bangla language,
disaster classification,
BanglaBERT,
ResNet50,
early fusion,
low-resource languages,
XLM-RoBERTa
\end{IEEEkeywords}
\vspace{0.1cm}
\IEEEpubidadjcol

\noindent\fbox{%
    \parbox{0.97\columnwidth}{%
        \footnotesize
        \textbf{IEEE Copyright Notice:} \copyright~2025 IEEE. Personal use of this material is permitted. Permission from IEEE must be obtained for all other uses, in any current or future media, including reprinting/republishing this material for advertising or promotional purposes, creating new collective works, for resale or redistribution to servers or lists, or reuse of any copyrighted component of this work in other works.
        
        \textbf{Publication:} Accepted for publication in IEEE SPICSCON 2025.
    }%
}

\section{Introduction}
Bangladesh frequently experiences natural hazards such as floods, cyclones, landslides, and fires with major effects on communities and economic stability. The increased prevalence and the intensity of such events underscore the pressing necessity for effective monitoring systems. Social networking sites have turned into valuable sources of live disaster news with citizens offering text-based descriptions of incidents and posting them frequently along with images that provide instant context~\cite{caragea2016informative}. This explosion of multimodal data offers a unique opportunity for automated disaster monitoring but extracting actionable insights from noisy, multilingual content is still a major technical challenge especially for low-resource languages such as Bangla~\cite{farjana2024gender}.

Traditional disaster monitoring systems rely on satellite imagery and official reports, which are vulnerable to delays and limited coverage. On the contrary, social media provides real-time, ground-level footage that supplements traditional data sources. Despite these promising developments, processing under-resourced languages like Bangla remains challenging due to the language's inherent complexity and the widespread use of informal expressions.

\begin{figure}[htbp]
\centering
\includegraphics[width=0.48\textwidth]{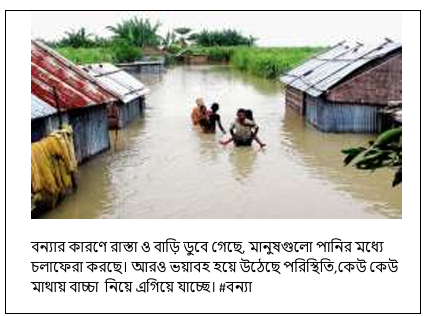}
\caption{Multimodal disaster content from social media.}
\label{fig:disaster_multimodal}
\end{figure}

Most of the research in disaster analytics so far has focused on unimodal techniques, processing text or image data independently. These techniques have reasonable performance for resourced languages, they do not capture the complementary character of multimodal information. Recent advances in deep learning have facilitated sophisticated multimodal fusion methods~\cite{kamoji2023fusion}. These methods are significantly under-explored for Bangla, and annotated multimodal datasets are scarce. Merging multimodal data has particular advantages in classifying disasters, as text provides clear explanations of events and outcomes while visual information offers immediate evidence of damage and context.

This paper introduces BanglaMM-Disaster, a comprehensive multimodal Bangla disaster classification framework. Our approach systematically incorporates transformer-based text encoders with convolutional neural networks for image analysis. The main contributions are:
\begin{itemize}
    \item We create a large-scale multimodal disaster dataset of 5,037 annotated social media posts covering nine disaster categories.
    \item We propose a transformer–CNN fusion framework that achieves 83.76\% accuracy and outperforms unimodal baselines.
    \item We conduct detailed error analysis, showing consistent cross-modal benefits to enable more reliable disaster understanding.
\end{itemize}

\section{Literature Review}
Automatic disaster analysis has moved from conventional rule-based systems to cutting-edge deep learning approaches. Early research focused on single-modality analysis, with separate streams for textual and visual content independently. This section reviews current methods, referring to trends for multimodal systems and identification of critical gaps in low-resource language processing.
\textbf{Unimodal Approaches for Disaster Analytics:}
Text-based disaster analysis has been extensively studied, particularly for high-resource languages. Caragea et al. \cite{caragea2016informative} were the first to utilize CNN for differentiating between informational disaster messages, while transformer models like BERT, Devlin et al. \cite{devlin2018bert} demonstrated better performance in contextual semantic comprehension. Recent work by Han et al. \cite{han2024quakebert} introduced QuakeBERT, achieving 84.33\% F1-score for earthquake disaster tweet classification, demonstrating the effectiveness of domain-specific transformer models. For image-based analysis, Kaur et al. \cite{kaur2022transfer} proved the effectiveness of transfer learning in hurricane damage detection, and Mouzannar et al. \cite{mouzannar2018damage} investigated crowdsourced photos for damage assessment.
\textbf{Multimodal Fusion Techniques: }
Recent multimodal disaster research has achieved significant breakthroughs through sophisticated fusion strategies. El-Niss et al. \cite{el2023multimodal} looked into federated learning approaches with F1-score 85.2\%, and Khattar et al. \cite{khattar2022camm} demonstrated cross-attention mechanisms with 84.08\% F1-score. Dar et al. \cite{dar2024crisisspot} presented CrisisSpot, a social context-aware graph-based multimodal framework achieving 88.1\% F1-score on the CrisisMMD benchmark, highlighting the importance of incorporating social context in disaster classification.
\textbf{Low-Resource Language Processing: }
Most of Bangla text analysis till date has focused on overall sentiment or abuse detection, rather than disaster events. Nabil et al. \cite{nabil2023bangla} collected Bangla social media content to categorize emergency posts, with 95.25\% F1 based on XLM-RoBERTa. Ghosh et al. \cite{ghosh2022gnom} mention about 84\% macro-F1 for multilingual disaster tweet classification using graph-augmented attention networks with transformers. Farjana et al. \cite{farjana2024gender} demonstrated that a CNN–LSTM model with BanglaBERT layer can reach 97.94\% accuracy for gender-abuse detection task in Bangla.
\textbf{Multimodal Research in Bangla:}  
Karim et al. \cite{karim2022multimodal} achieved 83\% F1-score using XLM-RoBERTa + DenseNet-161 for Bengali multimodal hate speech detection from memes and texts, demonstrating the viability of transformer-CNN fusion architectures for Bengali social media content analysis. Taheri et al. \cite{taheri2023bemofusionnet} obtained 77.5\% F1-score on emotion classification, while Alam et al. \cite{alam2024multimodal} addressed Bangla multimodal aggressive meme classification with 76\% weighted F1-score. However, both used datasets with fewer than 4,000 samples and did not include disaster-specific tasks.
\subsection{Research Gaps and Motivation}
Despite being used by over 300 million people in high-risk locations, there is currently no systematic multimodal framework for classifying disasters in Bangla. Second, Bangla multimodal studies have so far only been conducted on non-disaster applications and small-scale datasets. This paper addresses these shortcomings by offering the first thorough multimodal large-scale disaster classification system in Bangla.
\section{Dataset}
A novel multimodal dataset was constructed for Bangla disaster classification, comprising social media text and corresponding images capturing real-world disaster complexity in Bangladeshi scenarios.
\subsection{Data Collection and Annotation}
We gathered data from public Bangla social media platforms and local news websites focusing on disaster-related content. The collection process involved systematic sampling from Facebook posts, Twitter feeds, and online news portals during major disaster events in Bangladesh from 2020-2023. Two trained disaster-management annotators, native speakers of Bangla with expertise in emergency response, independently annotated all samples achieving Cohen's kappa ($\kappa = 0.82$), ensuring high mutual agreement. Our samples consist of single Bangla sentences and their corresponding images, annotated into one of nine disaster classes revealing disaster impacts in Bangladeshi scenarios.

\subsection{Dataset Statistics and Characteristics}
Our dataset includes 5,037 annotated samples well-balanced across different disaster classes. Text samples average 15.3 words with 12,847 distinct tokens, covering diverse linguistic patterns typical of Bangla social media communication. Image samples have standard size of 224×224 pixels and exhibit a broad range of visual contexts including infrastructure damage, natural landscapes, human activities, and weather conditions. We divided our dataset using strategic sampling: 70\% training (3,526 samples), 10\% validation (504 samples), and 20\% test (1,007 samples), ensuring balanced class distribution.
\begin{table}[htbp]
\centering
\caption{Class Distribution in the Bangla Disaster Dataset}
\label{tab:class_dist}
\begin{tabular}{lcc}
\hline
\textbf{Class} & \textbf{Count} & \textbf{Percentage} \\
\hline
Agricultural Damage (AD) & 800 & 16\% \\
Non Damage (ND) & 650 & 13\% \\
Infrastructural Damage (ID) & 450 & 9\% \\
Landslides (LS) & 400 & 8\% \\
Damage to Natural Landscape (DNL) & 600 & 12\% \\
Floods (FL) & 500 & 10\% \\
Fires (FR) & 300 & 6\% \\
Economic Loss (EL) & 400 & 8\% \\
Others (OT) & 937 & 18\% \\
\hline
Total & 5,037 & 100\% \\
\hline
\end{tabular}
\end{table}
\section{Methodology}
\begin{figure}[htbp]
    \centering
    \includegraphics[width=0.46\textwidth]{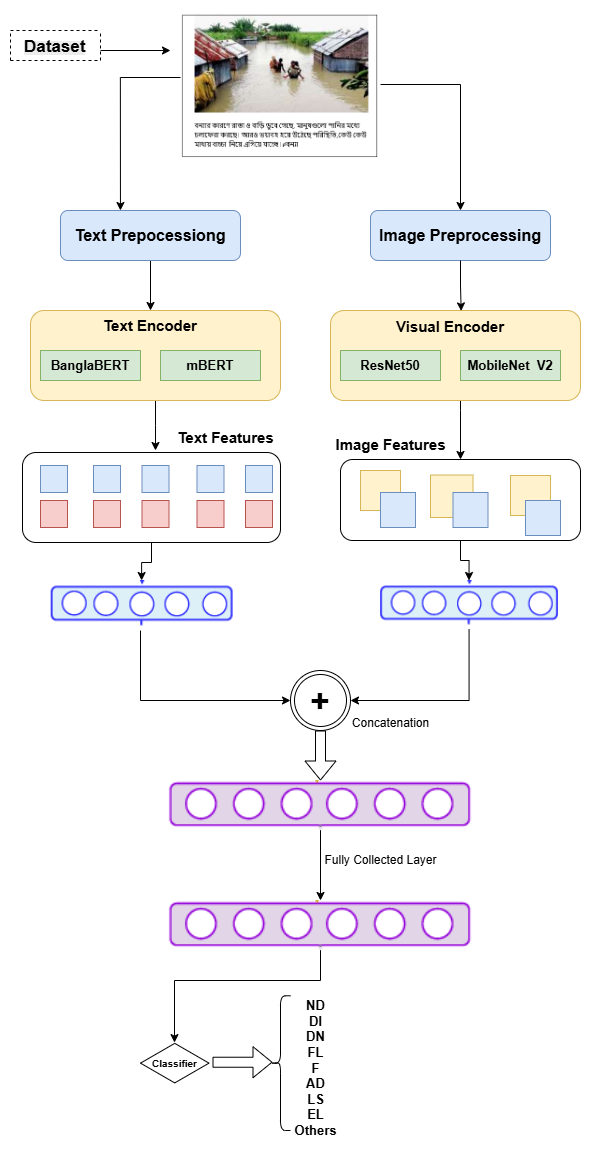}
    \caption{Overview of the proposed multimodal disaster classification framework.}
    \label{fig:methodology_overview}
\end{figure}

Our approach integrates Bangla text and image data by a single multimodal deep learning pipeline as in Figure~\ref{fig:methodology_overview}. The framework consists of four main components: data preprocessing, feature extraction, early fusion, and classification. The architectural choices were motivated by computational efficiency requirements for real-time disaster response and proven effectiveness in low-resource language scenarios. We selected BanglaBERT for native Bangla understanding, mBERT for multilingual robustness, and XLM-RoBERTa for superior cross-lingual transfer learning capabilities~\cite{bhattacharjee2022banglabert}. ResNet50, DenseNet169, and MobileNetV2 were chosen over Vision Transformers for computational efficiency in real-time disaster scenarios~\cite{dar2024crisisspot}.
\subsection{Data Preprocessing Pipeline}
The preprocessing involves making inputs of uniform format for both modalities.
\textbf{Text Preprocessing:} Input text undergoes deep cleaning by removing punctuation, unwanted whitespace, and non-textual aspects. English and Banglish text are translated into Bangla using Google Translate API, and frequent misspelling errors are corrected to achieve standard preprocessing for Bangla models. Special characters and emojis are normalized to preserve semantic meaning.
\textbf{Image Preprocessing:} All input images are rescaled to 224×224 pixels according to ImageNet standards, normalized to [0, 1] pixel values and standardized. Data augmentation techniques such as random horizontal flips, rotations (±15°), and zooms (0.8-1.2x) are applied to improve model generalization.
\subsection{Textual Feature Encoding} 
The text encoder processes Bangla disaster-related text using transformer-based models (BanglaBERT, mBERT, XLM-RoBERTa) following an effective approach to low-resource language understanding.
\textbf{Tokenization and Embedding:} All sentences are tokenized into sub-words based on WordPiece tokenization, handling out-of-vocabulary words characteristic of social media content. Tokens are embedded using positional encodings:
\begin{equation}
PE(pos, 2i) = \sin\left(\frac{pos}{10000^{2i/d_{model}}}\right)
\end{equation}
\begin{equation}
PE(pos, 2i+1) = \cos\left(\frac{pos}{10000^{2i/d_{model}}}\right)
\end{equation}
where $pos$ is the position, $i$ is the dimension index, and $d_{model}$ is the embedding dimension.
\textbf{Feature Extraction:} Contextual representations are produced by multi-head self-attention mechanisms:
\begin{equation}
\text{Attention}(Q, K, V) = \text{softmax}\left(\frac{QK^T}{\sqrt{d_k}}\right)V
\end{equation}
where $Q$, $K$, $V$ represent query, key, and value matrices, and $d_k$ is the key dimension. The final text representation is obtained from the [CLS] token embedding.
\subsection{Visual Feature Encoding}
Visual features of pre-trained CNNs (ResNet50, DenseNet169 and MobileNetV2) are obtained by eliminating top classification layers and adding global average pooling layers. These models trained on ImageNet offer stable feature representations for disaster images despite domain mismatch.
Hierarchical feature extraction computes the visual feature representation:
\begin{equation}
F_{visual} = \text{GlobalAvgPool}(\text{CNN}(I))
\end{equation}
where $I$ represents the input image and CNN denotes the convolutional feature extractor. This approach captures low-level patterns (textures, edges) and high-level semantic concepts (damaged buildings, flooding).
\subsection{Early Fusion and Classification}
We employ early fusion through feature concatenation over late fusion due to computational efficiency and superior performance for disaster classification tasks where textual and visual information are complementary~\cite{el2023multimodal}. The multimodal fusion concatenates textual and visual information:
\begin{equation}
F_{joint} = [F_{text}; F_{visual}] \in \mathbb{R}^{d_{text} + d_{visual}}
\end{equation}
where $d_{text} = 768$ and $d_{visual} = 2048$ for our best configuration. This early fusion concatenation enables cross-modal feature interactions at the representation level, allowing the model to learn joint patterns leveraging both semantic descriptions and visual evidence simultaneously.
This joint representation passes through fully connected layers with dropout (rate 0.1) for regularization, preventing overfitting on the limited disaster dataset. The final classification is performed using softmax activation:
\begin{equation}
\hat{y} = \text{softmax}(W_f F_{joint} + b_f)
\end{equation}
where $W_f \in \mathbb{R}^{C \times (d_{text} + d_{visual})}$ and $b_f \in \mathbb{R}^C$ are learned fusion parameters, and $C = 9$ is the number of disaster classes.
\subsection{Training Configuration}
The training process is designed to manage multimodal training with constrained disaster data. We use Adam optimizer with different learning rates: $1 \times 10^{-5}$ for pre-trained text encoders and $3 \times 10^{-5}$ for fusion layers respectively. Batch size is 32 with early stopping based on validation loss. The optimization follows categorical cross-entropy loss:
\begin{equation}
\mathcal{L} = -\sum_{i=1}^N \sum_{c=1}^C y_{ic} \log(\hat{y}_{ic})
\end{equation}
\section{Experimental Results}
This section validates comprehensive evaluation of the BanglaMM-Disaster framework comparing unimodal and multimodal approaches. We examine performance indicators, error patterns and show the capability of multimodal fusion on Bangla disaster classification.
\subsection{Unimodal Performance Analysis}
\subsubsection{Visual-Only Models}
\begin{figure}[htbp]
    \centering
    \includegraphics[width=0.47\textwidth]{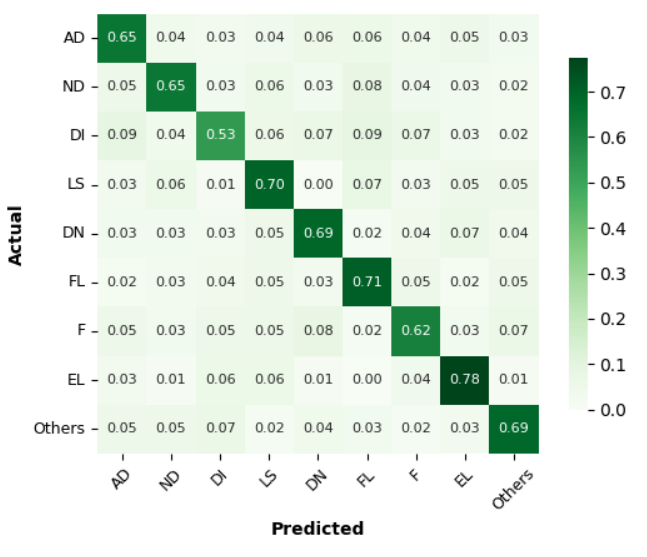}
    \caption{Confusion matrix for best visual model (ResNet50).}
    \label{fig:cm_visual}
\end{figure}

Table~\ref{tab:visual_results} summarizes image-only model performance. ResNet50 achieved the best accuracy of 66.85\%, outperforming DenseNet169 (65.47\%) and ResNet101 (64.32\%). Figure~\ref{fig:cm_visual} shows that visual models effectively separate distinct classes like Non Damage (ND) and Floods (FL), but struggle with ambiguous ones like Fires (FR) and Economic Loss (EL).
\begin{table}[htbp]
\centering
\caption{Performance of Image-Only Models}
\label{tab:visual_results}
\begin{tabular}{|l|c|c|c|c|}
\hline
\textbf{Model} & \textbf{Acc.} & \textbf{Prec.} & \textbf{Recall} & \textbf{F1} \\
\hline
ResNet50 & 66.85 & 66.62 & 67.08 & 66.85 \\
\hline
DenseNet169 & 65.47 & 65.23 & 65.71 & 65.47 \\
\hline
ResNet101 & 64.32 & 64.15 & 64.49 & 64.32 \\
\hline
MobileNetV2 & 62.41 & 62.17 & 62.65 & 62.41 \\
\hline
\end{tabular}
\end{table}
\subsubsection{Text-Only Models}
Text-only models performed significantly better than visual models as shown in Table~\ref{tab:text_results}. XLM-RoBERTa demonstrates superior multilingual support for Bangla text with 79.92\% accuracy. The model effectively distinguishes between Agricultural Damage (AD) and Non Damage (ND) but encounters challenges with textual complexity in Economic Loss (EL) and Others categories.
\begin{table}[htbp]
\centering
\caption{Performance of Text-Only Models}
\label{tab:text_results}
\begin{tabular}{|l|c|c|c|c|}
\hline
\textbf{Model} & \textbf{Acc.} & \textbf{Prec.} & \textbf{Recall} & \textbf{F1} \\
\hline
XLM-RoBERTa & 79.92 & 79.65 & 80.21 & 79.93 \\
\hline
mBERT & 78.15 & 77.82 & 78.48 & 78.15 \\
\hline
BanglaBERT & 76.73 & 76.29 & 77.18 & 76.73 \\
\hline
\end{tabular}
\end{table}
\begin{figure}[htbp]
    \centering
    \includegraphics[width=0.47\textwidth]{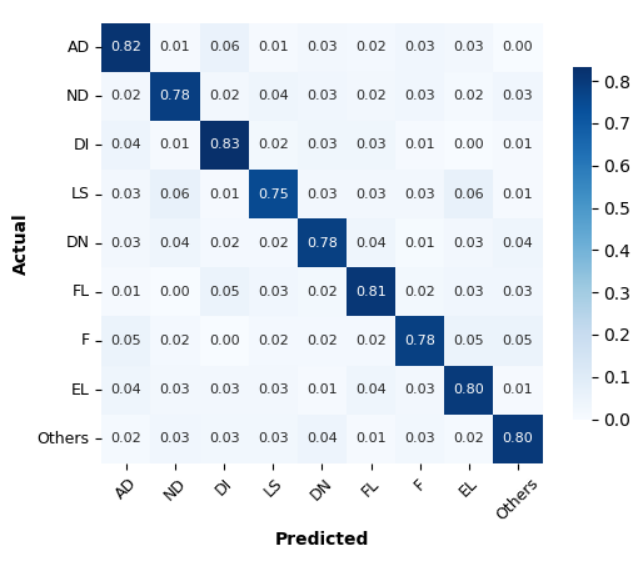}
    \caption{Confusion matrix for best text model (XLM-RoBERTa).}
    \label{fig:cm_text}
\end{figure}
\subsection{Multimodal Fusion Results}
Table~\ref{tab:multimodal_results} presents comprehensive multimodal performance. The optimized mBERT + ResNet50 fusion achieved the best accuracy of 83.76\%, representing a 3.84\% improvement over the best text-only model and 16.91\% over the best visual-only model. The early fusion concatenation mechanism enables this improvement by allowing cross-modal feature interactions that disambiguate ambiguous textual descriptions using visual confirmation. Even the worst multimodal combination outperformed the best visual-only baseline, demonstrating consistent multimodal improvements.
\renewcommand{\arraystretch}{1.3}
\setlength{\tabcolsep}{14pt}
\begin{table*}[htbp]
\centering
\caption{Performance of Multimodal Model Combinations}
\label{tab:multimodal_results}
{\small
\begin{tabular}{|l|c|c|c|c|}
\hline
\textbf{Model} & \textbf{Acc.} & \textbf{Prec.} & \textbf{Recall} & \textbf{F1} \\
\hline
\textbf{mBERT + ResNet50 (Optimized)} & \textbf{83.76} & \textbf{83.54} & \textbf{83.98} & \textbf{83.76} \\
\hline
mBERT + ResNet50 & 82.34 & 82.15 & 82.53 & 82.34 \\
\hline
XLM-RoBERTa + DenseNet169 & 81.87 & 81.64 & 82.11 & 81.87 \\
\hline
BanglaBERT + ResNet101 & 81.42 & 81.18 & 81.66 & 81.42 \\
\hline
mBERT + DenseNet201 & 80.95 & 80.73 & 81.17 & 80.95 \\
\hline
\end{tabular}
}
\end{table*}
\begin{figure}[htbp]
    \centering
    \includegraphics[width=0.47\textwidth]{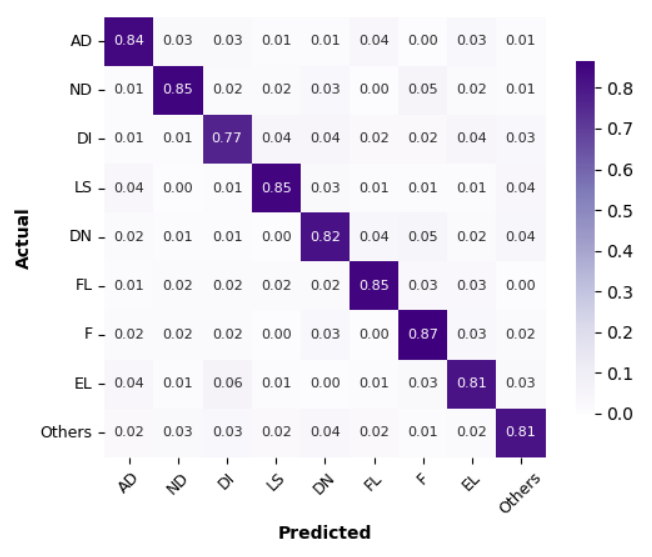}
    \caption{Confusion matrix for best multimodal model (mBERT+ResNet50).}
    \label{fig:cm_multimodal}
\end{figure}
\subsection{Comparative Analysis}
Table~\ref{tab:comparative_analysis} compares our results with related work. Due to the lack of Bangla multimodal disaster datasets, direct comparison is challenging as different datasets and languages are used across studies. However, our approach achieves competitive performance despite working with a morphologically complex low-resource language and challenging 9-class taxonomy. For real-time deployment, our framework achieves 0.45 seconds average inference time with 1.8GB memory footprint on standard GPU hardware.
\renewcommand{\arraystretch}{1.3}
\setlength{\tabcolsep}{6pt}
\begin{center}
\captionof{table}{Comparison with Related Work}
\label{tab:comparative_analysis}
{\footnotesize
\begin{tabular}{|l|l|l|c|}
\hline
\textbf{Method} & \textbf{Domain} & \textbf{Classes} & \textbf{F1 (\%)} \\
\hline
\multicolumn{4}{|c|}{\textit{Bengali Multimodal}} \\
\hline
Karim et al.~\cite{karim2022multimodal} & Hate Speech & 2 & 83.0 \\
\hline
Taheri et al.~\cite{taheri2023bemofusionnet} & Emotion & 4 & 77.5 \\
\hline
Alam et al.~\cite{alam2024multimodal} & Meme & 2 & 76.0 \\
\hline
\multicolumn{4}{|c|}{\textit{English Multimodal Disaster (Context)}} \\
\hline
CrisisSpot~\cite{dar2024crisisspot} & Disaster & 6 & 88.1 \\
\hline
El-Niss et al.~\cite{el2023multimodal} & Disaster & 6 & 85.2 \\
\hline
Khattar et al.~\cite{khattar2022camm} & Disaster & 5 & 84.1 \\
\hline
\multicolumn{4}{|c|}{\textit{Bengali Text-Only Disaster}} \\
\hline
Nabil et al.~\cite{nabil2023bangla} & Emergency & 2 & 95.3 \\
\hline
Ghosh et al.~\cite{ghosh2022gnom} & Disaster & Multi & 84.0 \\
\hline
\hline
\textbf{Our Work} & \textbf{Disaster} & \textbf{9} & \textbf{83.76} \\
\hline
\end{tabular}
}
\end{center}
\subsection{Error Analysis and Cross-Modal Benefits}
\begin{figure}[htbp]
    \centering
    \includegraphics[width=0.47\textwidth]{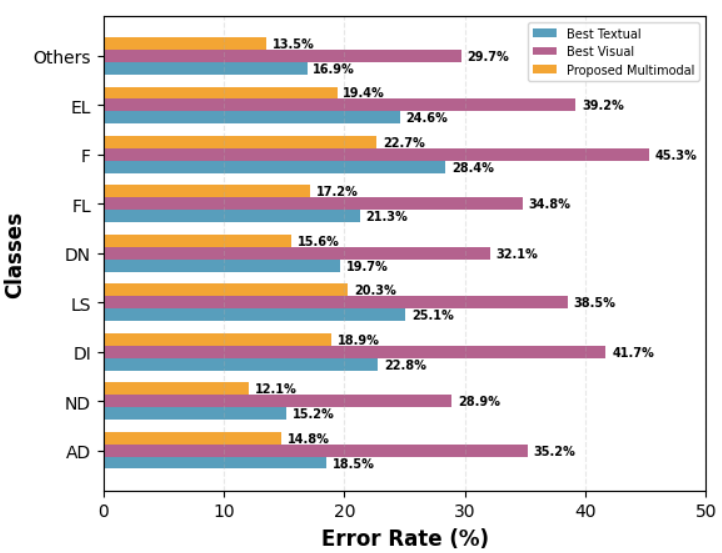}
    \caption{Class-wise error rates comparing best visual, text, and multimodal models.}
    \label{fig:error_rate}
\end{figure}

Figure~\ref{fig:error_rate} demonstrates consistent superiority of multimodal fusion across all disaster classes. The multimodal model achieves the lowest error rates for each disaster type, with complementary information from both modalities contributing to performance gains. For example, the Fires (FR) class error drops from 45.3\% (visual-only) and 28.4\% (text-only) to 22.7\% with multimodal fusion, a 50\% improvement over visual baseline. Similarly, Economic Loss (EL) errors decrease from 39.2\% (visual) and 24.6\% (text) to 19.4\% (multimodal). These consistent improvements demonstrate that the early fusion approach effectively captures cross-modal dependencies.
\section{Conclusion}
This research introduces BanglaMM-Disaster, a comprehensive multimodal deep learning framework for disaster classification in Bangla, addressing critical gaps in disaster analytics for low-resource languages. We developed a novel dataset of 5,037 annotated Bangla social media posts across 9 disaster categories and demonstrated that early fusion of transformer-based text encoders with CNN visual features achieves 83.76\% accuracy, with significant improvements of 3.84\% over text-only and 16.91\% over image-only baselines. Our error analysis confirms consistent cross-modal benefits across all disaster types. The framework's computational efficiency (0.45s inference, 1.8GB memory) enables practical deployment for real-time disaster monitoring systems. This work establishes new benchmarks for Bangla multimodal disaster classification and demonstrates the effectiveness of multimodal learning in low-resource linguistic settings.
Future directions include exploring attention-based fusion mechanisms and graph neural networks~\cite{dar2024crisisspot} to capture more complex cross-modal relationships for enhanced disaster understanding.
\bibliographystyle{IEEEtran} 
\bibliography{ref}
\end{document}